\documentclass[fleqn,10pt]{wlscirep}
\title{Novel digital tissue phenotypic signatures of distant metastasis in colorectal cancer}
\author[1,*]{Korsuk Sirinukunwattana}
\author[2]{David Snead}
\author[3]{David Epstein}
\author[4]{Zia Aftab}
\author[4]{Imaad Mujeeb}
\author[2]{Yee Wah Tsang}
\author[5]{Ian Cree}
\author[6]{Nasir Rajpoot}
\affil[1]{Institute of Biomedical Engineering, University of Oxford, UK}
\affil[2]{Department of Pathology, University Hospitals Coventry and Warwickshire, Coventry, UK}
\affil[3]{Mathematics Institute, University of Warwick, Coventry, UK}
\affil[4]{Hamad Medical Corporation, Doha, Qatar}
\affil[5]{International Association for Cancer Research, France}
\affil[6]{Department of Computer Science, University of Warwick, Coventry, UK}

\affil[*]{To whom correspondence should be addressed. Email: sirinukun.korsuk@gmail.com, n.m.rajpoot@warwick.ac.uk}


\usepackage{multirow}
\usepackage{url}

\begin{abstract}
Distant metastasis is the major cause of death in colorectal cancer (CRC). Patients at high risk of developing distant metastasis could benefit from appropriate adjuvant and follow-up treatments if stratified accurately at an early stage of the disease. Studies have increasingly recognized the role of diverse cellular components within the tumor microenvironment in the development and progression of CRC tumors. In this paper, we show that a new method of automated analysis of digitized images from colorectal cancer tissue slides can provide important estimates of distant metastasis-free survival (DMFS, the time before metastasis is first observed) on the basis of details of the microenvironment. Specifically, we determine what cell types are found in the vicinity of other cell types, and in what numbers, rather than concentrating exclusively on the cancerous cells. We then extract novel tissue phenotypic signatures using statistical measurements about tissue composition. Such signatures can underpin clinical decisions about the advisability of various types of adjuvant therapy.
\end{abstract}
\begin{document}

\flushbottom
\maketitle
\thispagestyle{empty}

%\noindent Please note: Abbreviations should be introduced at the first mention in the main text – no abbreviations lists. Suggested structure of main text (not enforced) is provided below.

\section*{Introduction}

Cell function and behavior cannot be fully understood without the context of their microenvironment. Communication between cells and their surroundings allows the functional organization of cells into tissues and organs. It also plays a vital role in maintaining tissue homeostasis by generating signals that suppress and revert malignant phenotypes \cite{bissell2011don}. Experiments in animal and cell culture models have demonstrated that certain conditions of the microenvironment can cause potent cancerous cells to revert to an almost normal phenotype \cite{stoker1990embryonic, weaver1997reversion}. Although the normal tissue microenvironment is known to be resilient to tumorigenesis, false signals in the microenvironment can disrupt tissue homeostasis and subsequently initiate tumors. The microenvironment in which tumor exists is both complex and heterogeneous, inhabited by a multitude of cellular and non-cellular components including tumor cells, extracellular matrix, tumor stroma, blood vessels, inflammatory cells, signaling molecules \cite{whiteside2008tumor,hanahan2011hallmarks,quail2013microenvironmental}. Studies over the last decade have increasingly recognized the role of these different components in the development and progression of tumors \cite{hanahan2011hallmarks}. This paper adds to this evidence, and shows how its quantification may be automated.

Metastasis is the major cause of morbidity and death in colorectal cancer (CRC). The 5-year survival rate in CRC patients with distant metastasis is approximately 10\%, considerably smaller than 70\% with regional metastasis and 90\% without metastasis \cite{siegel2016cancer}. Patients at high risk of developing distant metastasis could benefit from appropriate adjuvant and follow-up treatments if stratified accurately. The literature reports several histopathological features carrying prognostic value for CRC progression. Each of the features reflects competing cellular stimuli that influence tumor progression or suppression within the microenvironment. Type, density, and relative locations of different tissue components in the tumor microenvironment are crucial in determining progression and patient survival in CRC. For instance, the number of cytotoxic and memory T cells in the tumor center and the invasive margin have been linked to an improved prognosis of CRC \cite{galon2006type}. Similarly, numerous studies have reported cancer-associated fibroblasts (CAFs) and desmoplasia to be important histopathological features associated with an unfavorable prognosis for CRC and an increased mortality rate \cite{mesker2007carcinoma,crispino2008role,west2010proportion,huijbers2012proportion,jayasinghe2015histopathological}. Analogous to a wound that never heals \cite{dvorak1986tumors,schafer2008cancer}, tumors stimulate many associated responses, wherein normal fibroblasts have been reported to acquire a cancer-associated phenotype (5,16). Furthermore, the extent of necrosis in CRC has been reported to correlate strongly with cancer progression and patient survival \cite{jayasinghe2015histopathological,pollheimer2010tumor,richards2012prognostic}. The link between necrosis and tumor progression is possibly due to the hypoxic nature of tumors, which drives tumor infiltrating inflammatory cells, namely phagocytic macrophages and granulocytes, to secrete pro-inflammatory cytokines which in turn promote cell proliferation \cite{whiteside2008tumor}.
	
In this study, we investigate the significance of tissue phenotypic and morphometric features, exploring in particular cellular heterogeneity in tumor microenvironments, in determining metastatic potential in CRC patients diagnosed with advanced primary tumors. Based on the AJUCC/UICC-TNM staging system \cite{edge2010american}, this group of patients have a primary tumor that has grown through the outer lining of colon wall (T3/T4), have no lymph nodes that are affected by cancer cells (N0), and may or may not have distant metastasis at the time of diagnosis (M0/M1). Detailed quantitative analysis was performed on whole slide images (WSIs) of CRC histology slides, stained with routine Hematoxylin \& Eosin (H\&E) dyes in a fully quantitative manner, using bespoke image analysis methods to provide an objective and reproducible assessment. Quantitative analysis of various types of cell population reveals novel tissue phenotypic features, derived from both cell-cell connection frequencies and tissue appearance, with significant association with metastasis incidence and distant metastasis-free survival (DMFS) in the advanced primary CRC tumors.

\section*{Results}
\subsection*{Quantifying tissue phenotypic signatures of CRC tumors}
In this study, WSIs of Hematoxylin and Eosin (H\&E)-stained histological sections from 108 patients with advanced node negative primary CRC tumors (T3/T4, N0, M0/M1) were acquired from two independent cohorts from two different institutes: University Hospitals Coventry and Warwickshire (UHCW, 78 patients) and Hamad General Hospital (HGH, 30 patients). Summary details of the cohorts and clinical information are given in Table \ref{tab:clinical_data}. 

CRC, like other solid tumors, is a disease of substantial heterogeneity \cite{punt2017tumour,dalerba2011single}. Different parts of the same tumor can exhibit different features including cellular morphology, gene expression, metabolism, motility, angiogenic, proliferative, immunogenic, and metastatic potential \cite{marusyk2010tumor}. The tumor microenvironment is composed of diverse cell types; each plays a different role in tumor development and progression — some support and promote tumor progression while others play host protective roles \cite{hanahan2011hallmarks}. The biological functions of cells are not only determined by their type but are also greatly influenced by their surrounding context.  It follows that tissue morphometric signatures measuring tumor heterogeneity could be computed from the analysis of distributions and relative locations of cellular populations in the tumor microenvironment.

\begin{figure}[ht]
\centering
\includegraphics[width=0.6\textwidth]{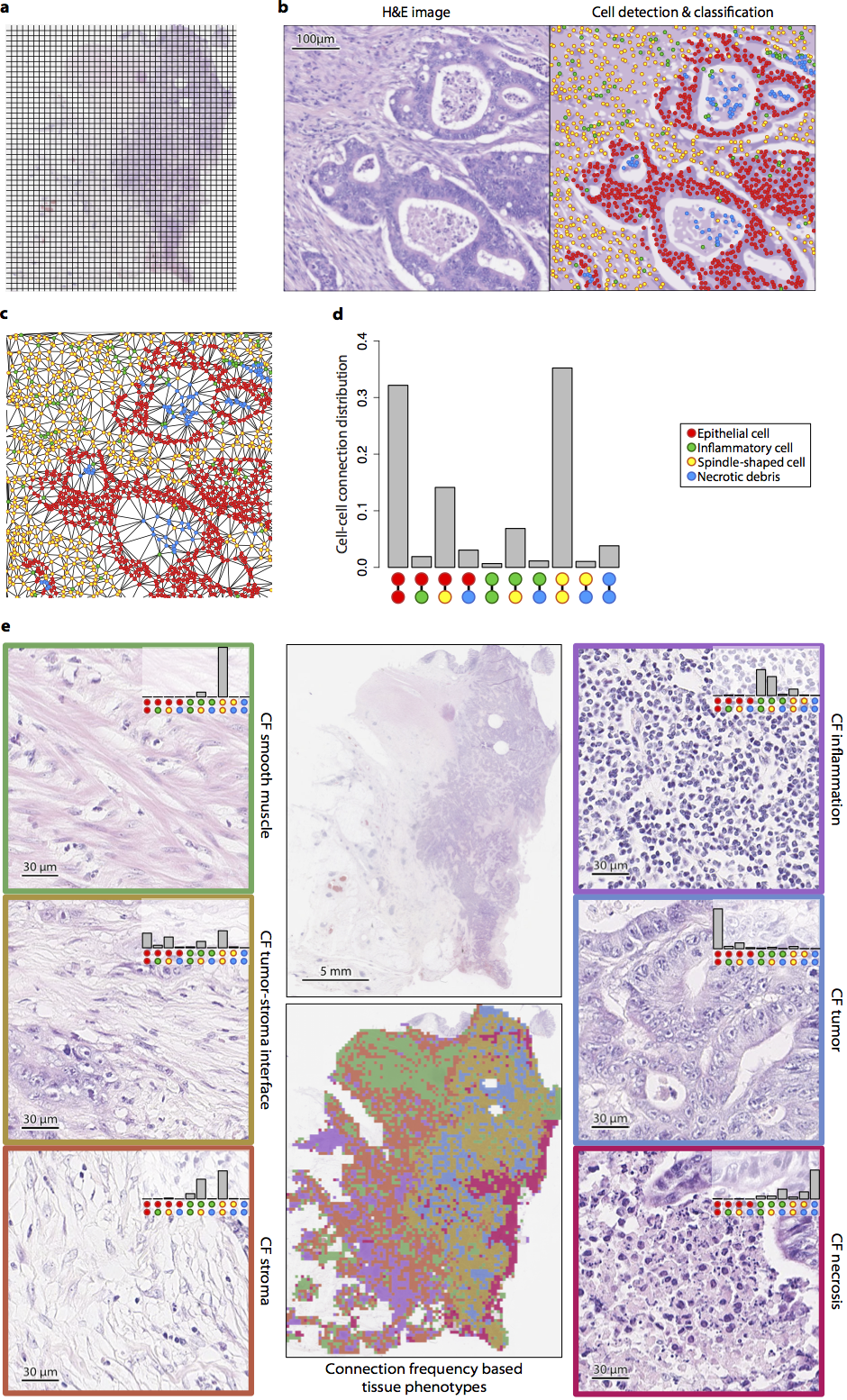}
\caption{Profiling tissue morphometric phenotypes. A WSI was divided into small regions of size 200$\times$200 $\mathrm{\mu m^2}$ (a). Cellular components in the image were localized and classified into 4 different cell types, including malignant epithelial cell, inflammatory cell, spindle-shaped cell, and necrotic debris, based on their nuclear morphology and surrounding tissue context (b). A cell network was subsequently constructed from the cell detection and classification results, in which nodes in the network represent cells and edges conceptualize relationships among them (c). A distribution of cell-cell connections was calculated for each small region (d). According to their distributions of cell-cell connections, tissue regions were profiled into 6 different phenotypes (e).}
\label{fig:overview}
\end{figure}

Here, we outline the quantification of digital tissue phenotypic signatures (see Methods for details). We divided each tumor histology image (i.e., each WSI) into small square regions or sub-images (Fig. \ref{fig:overview}a) and analyzed the small sub-images to obtain local characteristics that were then summarized to characterize the entire tumor section. We first applied our artificial intelligence (AI) based algorithm \cite{sirinukunwattana2016locality}, which was recently shown to be the state-of-the-art in detecting and distinguishing between four types of cells based on their morphology and context, to each sub-image. The four types of cells were: malignant epithelial cells, spindle-shaped cells (normal fibroblasts, cancer-associated fibroblasts and smooth muscle cells), inflammatory cells (eosinophils, lymphocytes and neutrophils), and necrotic debris (Fig. \ref{fig:overview}b). This allowed us to do quantification of tissue morphological characteristics associated with tumor, based on both distributions and relative spatial locations of diverse cell types. For each small tissue region (sub-image) in the large WSI, we then constructed a cell network (Fig. \ref{fig:overview}c).  Each vertex of the network represents a cell of a certain type, and an edge denotes a cell-cell connection between immediately neighboring cells. Based on the distribution of cell-cell connections in the network (Fig. \ref{fig:overview}d), we then grouped the local tissue regions into different phenotypes using an unsupervised learning approach. The six resulting connection frequency (CF) based tissue phenotypes were visually discernible with each phenotype corresponding mainly to local areas of smooth muscle, inflammation, tumor-stroma interface, tumor, stroma, or necrosis (Fig. \ref{fig:overview}e). Finally, we used the ratio of the area of each CF tissue phenotype to the total tissue area to give digital tissue phenotypic signature of each tumor sample (Methods).

To further examine the extent to which the aforementioned automatically derived cell-cell CF tissue phenotypes correlate with known tissue types, we also quantified the tissue types by means of appearance based (AP) tissue segmentation. The tissue content of each WSI was automatically segmented into the following eight categories: tumor, stroma, loose connective tissue, normal/hyperplastic mucosa, smooth muscle, necrosis, fat, and inflammation (Fig. S1). We then investigated correlation between the CF and AP based tissue phenotypes. These are smooth muscle, inflammation, tumor, stroma, and necrosis. The Spearman correlation coefficients for individual pairs of CF and AP features range from 0.427 to 0.698 (Fig. S2), indicating moderate correspondence between the automatically-derived phenotypes and the underlying tissue types.

In addition to the phenotypic and standard clinical features, we considered the following automatically-derived features: Morisita index \cite{maley2015ecological}, stroma-tumor ratio \cite{mesker2007carcinoma,west2010proportion,huijbers2012proportion}, and necrosis-tumor ratio \cite{jayasinghe2015histopathological,pollheimer2010tumor,richards2012prognostic}. These features have previously been identified as having prognostic significance for CRC or other malignancies. Morisita index measures the spatial coexistence of inflammatory cell and malignant epithelial cells \cite{maley2015ecological}. Stroma-tumor ratio is defined as the proportion of the total area of stroma to the total area of combined stroma and tumor in the tissue. Necrosis-tumor ratio is defined in the similar manner as that of stroma-tumor ratio (Methods). It is worth noting that in the above studies, the stroma-tumor ratio and necrosis-tumor ratio were semi-quantitatively assessed on manually selected small regions of histological slides. In contrast, we measured these quantities with greater precision and using all regions of our WSIs, thus avoiding subjective bias.

\subsection*{Association between phenotypic and clinical features}
Here, we determined the strength of association between the CF tissue phenotypic features and standard clinical features normally used in routine prognostication of colon cancer (Table \ref{tab:association}). The clinical features included tumor differentiation, tumor histological type, and primary tumor (T) stage. For example, to check whether there is  association between the CF inflamation ratio and the T stage, we test if the distribution of CF inflammation ratio of the group of samples that are annotated as pT3 stage is significantly different from the distribution of samples that are annotated as pT4 stage using Mann-Whitney U test (also known as Wilcoxon rank-sum test).

We found statistically significant association between CF inflammation ratio and T stage (p-value = 0.002) and between CF tumor ratio and tumor differentiation (p-value = 0.033). Nonetheless, the relatively small values of the coefficients of determination (CF inflammation ratio vs. T stage: $r^2$ = 0.092; CF tumor ratio vs. differentiation: $r^2$ = 0.042) indicate that CF inflammation ratio and CF tumor ratio are only weakly associated with T stage and tumor differentiation, respectively. There is no statistically significant association between other pairs of the CF phenotypic features and the standard clinical features. Altogether, these results suggest that the CF tissue phenotypic features are not strongly associated with standard clinical features and, therefore, are potentially new features whose prognostic significance is worth further investigation. 

\subsection*{Logistic regression analysis}
To assess the significance of each phenotypic feature in identifying a patient’s risk of having distant metastasis at the time of diagnosis or future progression, we carried out logistic regression analysis. Odds ratio factor and 95\% confidence interval (CI) estimates were obtained for each feature to quantify the risk of distant metastasis incidence associated with the phenotypic features (Methods). 

The results show that CF smooth muscle and inflammation ratios are statistically significant (p-value $<$ 0.05) in univariate analysis (Table \ref{tab:logistic}). Moreover, when adjusted for the effects of the standard clinical parameters in multivariate analysis, CF smooth muscle and inflammation ratios are statistically significant features (p-value $<$ 0.05), independent of other standard clinical features (Table \ref{tab:logistic} and Table S1). The interquartile change in CF smooth muscle ratio increases the odds of distant metastasis by a factor of 2.258 (95\% CI: 1.117 - 4.565) in univariate analysis and by 2.350 (95\%CI: 1.132 - 4.876) in multivariate analysis. The interquartile change in CF inflammation ratio, on the other hand, decreases the odds by a factor of 0.279 (95\% CI: 0.119 - 0.656) in the univariate analysis and 0.307 (95\%CI: 0.126 - 0.749). Despite the fact that CF smooth muscle and inflammation ratios are separately shown to be statistically significant in both the univariate and multivariate analyses, when considered together in the multivariate model (Table S1), their joint contribution towards the prediction of metastasis development becomes less clear. This is likely due to a moderate degree of correlation ($\rho$=-0.652) between the features. Thus, when one is used, the other should probably be disregarded.

Next, we investigated if the above statistical results could be achieved by means of the AP smooth muscle ratio and AP inflammation ratio features. Only the AP inflammation ratio is shown to be statistically significant in the univariate analysis (p-value = 0.027, Table \ref{tab:logistic}) and marginally significant in the multivariate analysis (p-value = 0.056, Table \ref{tab:logistic}). %\textcolor{red}{This piece of evidence supports the prognostic value of CF inflammation ratio, but at the same time does not rule out the plausibility of CF smooth muscle ratio, since the CF phenotypic features correlate moderately with the AP phenotypic features.}

\subsection*{Distant metastasis-free survival analysis}
Next, we investigated the prognostic significance of various features, using DMFS as a criterion. The analysis was carried out on all cases from the UHCW cohort (78 cases), for which survival data were available. In our multivariate analysis, the effect of individual features was adjusted for the effect of standard clinical features.

The tissue CF phenotypic features (smooth muscle, inflammation, and stroma ratios) and the AP phenotypic feature (inflammation ratio) were shown to be influencing features in determining the DMFS probability of the patients under Cox proportional hazards models (p $<$ 0.05, Table \ref{tab:cox}, and Table S2). In particular, the effect of the interquartile change in CF smooth muscle ratio is to increase the hazard by 2.138 times (95\% CI: 0.188 – 0.388) in the univariate analysis and by 2.467 times (95\% CI: 1.062 – 5.73) in the multivariate analysis. The interquartile change in AP inflammation ratio affects the DMFS probability by reducing the hazard by a factor of 0.401 (95\% CI: 0.026 - 0.075) in the univariate analysis and by a factor of 0.431 (95\% CI: 0.224 - 0.832) in the multivariate analysis. In addition, when CF smooth muscle and AP inflammation are compared together in the same multivariate model, the effects of AP inflammation ratios on the DMFS probability vanishes (Table S2). This is likely due to the correlation between these features ($\rho$ = -0.252).

The interquartile change in CF inflammation ratio influences the survival probability by decreasing the hazard by a factor of 0.412 (95\% CI: 0.204 - 0.934) in the univariate analysis and by a factor of 0.413 (95\% CI: 0.201 – 0.849) in the multivariate analysis. However, the effect of CF inflammation ratio is only marginally significant (p-value = 0.051) when adjusted by the standard clinical parameters in the multivariate analysis. Interestingly, CF stroma ratio also shows up as statistically significant in the univariate analysis (p-value = 0.032, hazard ratio factor = 0.475, 95\% CI: 0.252 - 0.896).

\begin{figure}[ht]
\centering
\includegraphics[width=0.5\textwidth]{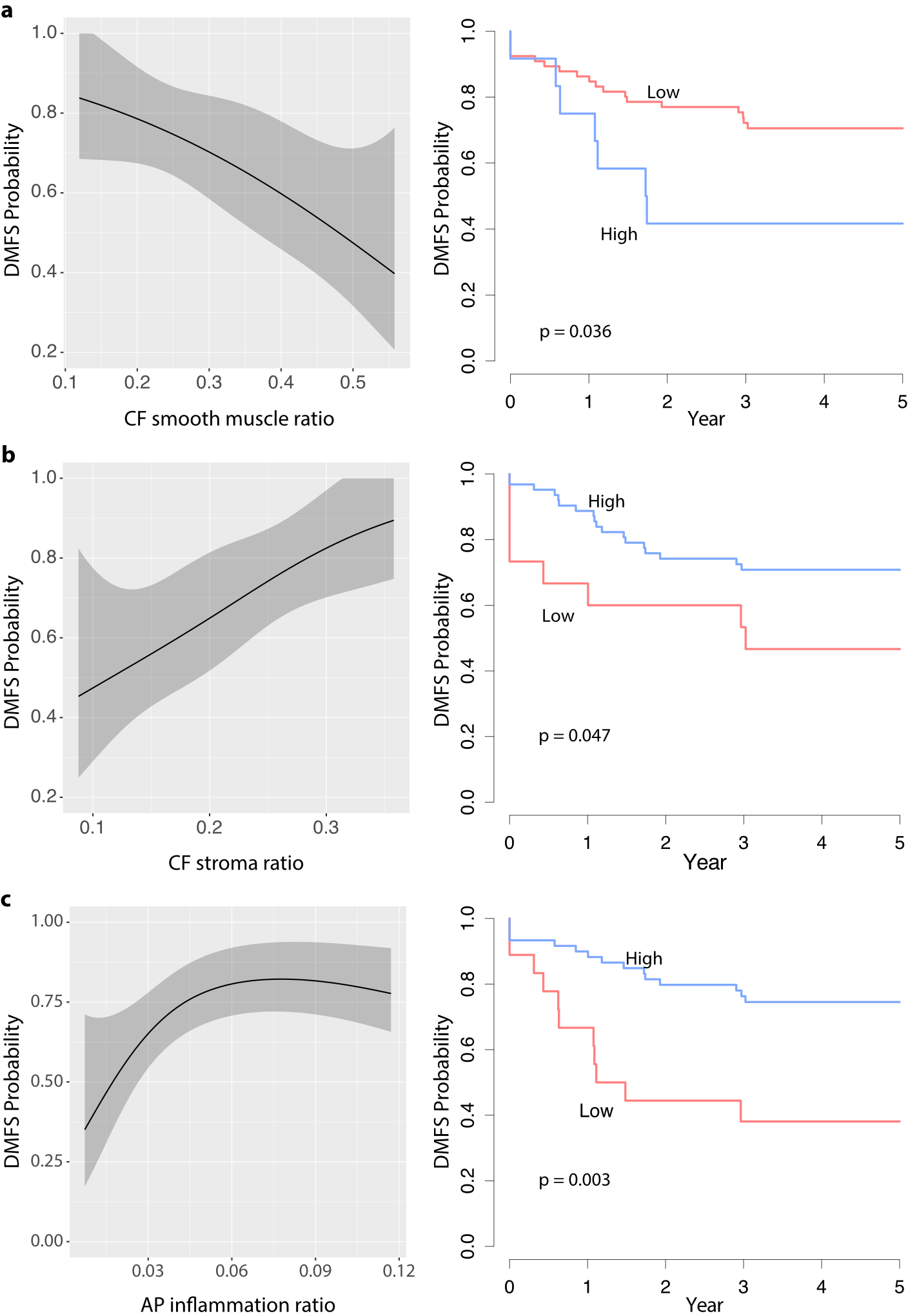}
\caption{Prognostic values of tissue phenotypes in the univariate survival analysis. (Left) A 5-year DMFS estimate with respect to the CF smooth muscle ratio (a), CF stroma ratio (b), and AP inflammation ratio (c). The gray shaded regions indicate the 95\% confidence intervals of the estimates. (Right) Kaplan-Meier curves stratified by the CF smooth ratio (a), CF stroma ratio (b), and AP inflammation ratio (c). A log-rank p-value was computed for each pair of Kaplan-Meier estimates.}
\label{fig:survival}
\end{figure}

We found that lower values of CF smooth muscle ratio and higher values of CF inflammation, CF stroma, and AP inflammation ratios were associated with low 5-year survival probabilities (Fig. \ref{fig:survival} left and Fig. S3d left). Furthermore, there are statistically significant differences between the survival distributions of cases when stratified by CF smooth muscle, CF stroma, and AP inflammation ratios (p-value $<$ 0.05, Table \ref{tab:cox}, Fig. \ref{fig:survival} right). Stratification by other features does not yield statistically significant results (Fig. S3).

In summary, CF smooth muscle and AP inflammation ratios are consistently shown to be important prognostic factors for DMFS across three different types of survival analyses, including univariate Cox regression analysis, multivariate Cox analysis, and log-rank test. Nonetheless, they are not shown to be independent of each other and therefore when one is used, the other should probably be disregarded.

\section*{Discussion}

The goal of this study was to investigate the prognostic significance of novel image-based quantitative morphometric features derived from diverse cellular populations that constitute the tumor microenvironment of CRC with advanced primary tumors (T3/T4, N0, M0/M1). 

\paragraph{Digital Phenotypic Features vs Histological Features.} To fully explore the rich microscopic level information available in a tissue section, we have developed an automated system to provide quantitative measurements and avoid bias due to observer variability. The analysis was conducted on WSIs of H\&E-stained formalin-fixed paraffin-embedded (FFPE) histological sections. Unlike previous works that identify diverse cellular components in a tumor section \cite{yuan2012quantitative,nawaz2015beyond}, our morphometric features are not limited to tumor cells, lymphocytes, and stromal cells, but also include other types of inflammatory cells, spindle-shaped cells, and necrotic debris. In addition, we explored the relationship between these cellular components through a cell-cell network in order to characterize the morphological and tissue phenotypic heterogeneity of tumor. Our system did not adopt a commonly used approach \cite{beck2011systematic,yu2016predicting} that calculates a large number of features followed by feature selection methods to select a handful of features suitable for the objectives of the analysis. Although such an exploratory approach has proved successful in some applications \cite{beck2011systematic,yu2016predicting}, the resulting features may not be easily interpretable in clinical terms. Moreover, if sufficiently many features are tried, it is likely that one of them will turn out to be ``statistically significant'' and so this approach requires follow-up tests of reproducibility.  Instead, we investigated a small set of meaningful quantitative features, automatically found through unsupervised phenotyping and segmentation.

Our systematic analysis shows that (a) the CF smooth muscle and CF inflammation ratios are potentially independent markers predicting the occurrence of distance metastasis (binary logistic regression analysis) and (b) the CF smooth muscle and AP inflammation ratios are potential prognostic markers affecting 5-year DMFS for CRC patients diagnosed with advanced primary tumor (Cox proportional hazards regression analysis). CF smooth muscle ratio essentially measures the amount of the smooth muscle that is part of the colon wall.  It quantifies the extent of spread and potential advancement of the tumors —  the concept is related (but not similar) to other measures such as T stage, tumor-stroma ratio \cite{mesker2007carcinoma,west2010proportion,huijbers2012proportion}, and tumor border configuration \cite{koelzer2014tumor,karamitopoulou2015tumour}. Low CF smooth muscle ratio is strongly associated with favorable prognosis. CF inflammation and AP inflammation ratios largely measure the amount of inflammation within the tumor tissue. High inflammation ratio is strongly associated with favorable prognosis, which supports the host-protective role of inflammatory cells in CRC that has been described by several studies \cite{galon2006type,pages2010immune,ohtani2007focus}. From this observation, one may hypothesize about the biological relevance of each of our automatically derived tissue phenotypes for tumor development and progression.

The prognostic value of stroma-tumor ratio \cite{mesker2007carcinoma,west2010proportion,huijbers2012proportion} and necrosis-tumor ratio \cite{jayasinghe2015histopathological,pollheimer2010tumor,richards2012prognostic} could not be confirmed in this study. It should be emphasized that, in those studies, both the ratios were semi-quantitatively measured in manually selected tumor-rich areas and were inevitably prone to observer bias. By contrast, our study measured these quantities in a fully automated and quantitative manner from all regions of the tumor section and therefore can be considered to be more objective and reproducible.

Uncertainty found in our analysis pertaining to the prognostic impact of standard clinical factors has also been confirmed in existing literature \cite{jass1986grading,compton1999pathology,hyngstrom2012clinicopathology,kim2013prognostic,zeng1992serosal,shepherd1997prognostic}. Despite the fact that tumor differentiation has been consistently shown to be a prognostic feature independent of stage \cite{chapuis1985multivariate,griffin1987predictors,wiggers1988regression,newland1994pathologic,jessup1998national}, the conventional grading process is subjective by its very nature and can exhibit a substantial degree of observer variability \cite{jass1986grading,compton1999pathology}. Inconsistency in grading is likely to occur in this study since samples from different cohorts were graded by different pathologists (UHCW: DS, HGH: IM). It is also worth noting that according to the revised WHO criteria \cite{hamilton2000classification}, only poorly differentiated tumor histology without mismatch repair protein deficiency is considered a high-risk factor. Presence of the mucinous histologic type in general is not an independent prognostic factor, given that available results are contradictory \cite{hyngstrom2012clinicopathology,kim2013prognostic}. Recent data have demonstrated the primary tumor extent (T4 stage) to be a likely prognostic factor for recurrence/metastasis \cite{kerr2009quantitative,salazar2010gene,tsikitis2014predictors}. Nevertheless, like other semi-quantitative features, there have been reports of variability in assessment of the degree of tumor extent \cite{zeng1992serosal,shepherd1997prognostic}. Results from our analysis also indicate that T4 tumors have adverse DMFS outcome compared to T3 tumors, though the difference is not statistically significant (p-value = 0.156). 

The majority of samples in this study come from patients diagnosed with stage II (Dukes stage B) CRC. This is characterized by advanced primary tumor with neither lymph node nor distant metastasis involvement (T3/T4, N0, M0). Stage II CRC consists of a heterogeneous population; some subgroups appear more likely to develop distant metastasis than others. Although adjuvant chemotherapy treatment is effective in other stages of the disease, there is a limited incremental benefit that stage II CRC patients could derive from this type of treatment in general \cite{hartung2005adjuvant,quasar2007adjuvant,schippinger2007prospective}. Due to the high financial cost and morbidity of the treatment coupled with uncertainty over which patients will relapse, there has long been a debate as to whether adjuvant chemotherapy treatment should be given to the patients, since a majority of the patients will already have been cured by surgical resection alone. In the absence of molecular or genetic predictive markers for chemotherapy response \cite{kerr2009quantitative,ribic2003tumor,kim2007prognostic}, improved prognostication accuracy seems to be the only key to better identify candidates who could potentially benefit the most from systemic therapies and thereby avoid unnecessary overtreatment as well as provide more efficient use of healthcare resources.

Even though several histological features have been demonstrated to be potential prognostic markers for recurrence or distant metastasis in stage II CRC, their prognostic significance is less clear and needs further validation. Primary tumor (T) stage and the number of lymph nodes examined have been recommended as risk factors by the National Comprehensive Cancer Network \cite{national2010nccn}. However, the results from our analysis do not support the T stage as a risk factor. Moreover, there is a controversy as to whether examining more lymph nodes can, in fact, reduce tumor staging error and in turn result in improved stage II patient survival \cite{wong2007hospital,moore2010staging}. High-frequency microsatellite instability has been associated with improved disease-free survival in one study \cite{kim2007prognostic} while in another study the effect was the opposite \cite{ribic2003tumor}. Gene expression profile is another factor that has shown promise for prediction of recurrence \cite{kerr2009quantitative,salazar2010gene}.

\paragraph{Study Limitations.} Based on the makeup of our dataset and the results from our analysis, we hypothesize that high CF smooth muscle ratio and low CF or AP inflammation ratios are potential risk factors for distant metastasis in stage II CRC. There are nevertheless some limitations of this study as described below.

Firstly, although our cell detection and classification approach \cite{sirinukunwattana2016locality} was developed to be robust to a certain degree of variation of images arising from factors such as stain inconstancy, batch effects, failed autofocus, and artefacts in the tissue preparation process, it remains to be tested if the degree of variation is excessive. Good image quality is therefore critical if the system is to produce accurate results. This issue can be addressed by careful tissue preparation and slide scanning.

Secondly, due to the nature of the H\&E stain and cellular morphology, our system is capable of identifying only a limited number of cell categories that are somewhat coarse. IHC stains could provide an effective means of identifying more specific cell types, such as different types of immune cells and fibroblasts (normal fibroblasts or CAFs), at the additional costs of IHC slide preparation and associated antibodies. 

Thirdly, the phenotyping proposed in this work was done on the basis of local cell-cell connection frequencies and also on the basis of appearance and other important contextual information such as tissue textures. This, on the one hand, can be seen as a limitation of the proposed quantitative tissue phenotyping approach, as it relies on local cell populations to generate global statistics. On the other hand, a number of studies have reported that normal cells of various types undergo transformation when coming into contact with tumor cells, thus resulting in some of the previously normal cells exhibiting new biological functions different from the original ones. The proposed approach focuses on cellular morphology and cellular context and avoids influences from other possibly misleading contextual information.

Finally, our analysis was based on a single dataset consisting of two independent cohorts from different institutes. To further confirm the reproducibility of the results and generalizability of our automated histologic quantification system, large-scale validation using independent cohorts from multiple institutes is required. To be translated into clinical practice, these limitations will need to be carefully addressed.

\paragraph{The Outlook.} With the increasing uptake of digital slide scanning technology in histopathology laboratories, digitized WSIs will gradually replace glass slides in routine pathology workflow \cite{snead2016validation}. This presents an opportunity to advance image analytical techniques and computational algorithms for quantitative analysis of tissue morphology and consequently to provide an accurate and reproducible means for the diagnosis and prognostication of cancers. This is the first step towards effective treatment, decision-making, and personalized medicine with computational support. In this work, we have demonstrated the usefulness of such morphometric tools to reveal prognostic features in CRC. Our morphometric analysis is not restricted to images of FFPE CRC tissues but is also applicable to frozen tissue images as well as to images from different types of cancers. This morphometric approach was not designed to replace pathologists, but rather to provide additional information to assist in their diagnostic decision-making and risk stratification. Another potentially important direction would be to investigate potential associations between genomic alterations and digital tissue phenotypic signatures reflecting measurable aspects of in the tumor microenvironment.

\section*{Methods}
\subsection*{Experimental Design}
The main objective of this study was to assess the significance of tissue phenotypic features for determining distant metastasis in advanced primary CRC. Specifically, we asked what quantitative tissue phenotypic features are biologically meaningful and important in predicting the concurrence at the time of diagnosis or subsequent development of distant metastasis and the distant-metastasis-free survival. Based on results from our statistical analyses, we have shown that digital tissue phenotypic features are independent prognostic factors for distant metastatic potential in CRC patients with advanced primary tumors (T3/T4, N0, M0/M1). The sample size for logistic and Cox proportional hazards regression analyses was calculated based on the concept of events per variable \cite{concato1995importance,peduzzi1995importance,peduzzi1996simulation} which indicates that a minimum of 30 metastatic subjects would be sufficient to control for a type I error rate at 7\%, 95\% CI coverage of 93\% , and a relative bias of 7\% of the estimate in the Wald test. We retrospectively recruited CRC subjects with advanced primary tumors. Cases without a 5-year distant metastasis status were excluded and the enrollment was stopped when the calculated sample size was reached. Our analyses were conducted on H\&E-stained WSIs of tumor sections. After reviewing the WSIs, we further excluded outlier cases whose tissue section had no tumor. In view of the limited number of cases, randomization was not used in any experiments. 

\subsection*{Patient and clinical information}
This study involved two independent cohorts of CRC patients from two institutes. The first cohort consisted of 78 patients initially admitted for CRC treatment during the years 2006 to 2010 at University Hospitals Coventry and Warwickshire (UHCW), Coventry, UK. The second cohort comprised 30 patients admitted during the years 2007 to 2012 at Hamad General Hospital (HGH), Doha, Qatar. For each case, clinical data included tumor histological type, differentiation, stage of the primary tumor (T), lymph node metastasis (N), and distant metastasis (M). The 5-year DMFS data were available only for UHCW cases. All CRC patients were diagnosed with locally advanced tumors (T3/T4) and negative lymph node (N0). Some patients appeared to have distant metastatic tumors at the time of diagnosis (M1), but the majority did not (M0). The TNM classification was reviewed and conducted according to the AJUCC/UICC-TNM staging system \cite{edge2010american}. Summary details of the clinical information are given in Table \ref{tab:clinical_data}. 

The data used for this study including the WSIs and clinical information was provided after de-identification and informed patient consent was obtained from all subjects. Ethics approval for this study was obtained from the National Research Ethics Service North West (REC reference 15/NW/0843). All the experiments were carried out in accordance with approved guidelines and regulations.

%Most of the two cohorts did not receive neoadjuvant chemotherapy or radiotherapy before surgery. 

\subsection*{Histological samples and Imaging}
For each case, tissue sections were prepared from an FFPE tumor tissue block and were then stained with H\&E. Each tissue section was prepared in the pathology laboratory of the UHCW hospital.  Histological slides were digitally scanned using the Omnyx VL120 Scanner (GE Omnyx, LLC) with an $\times$40 setting (equivalent to 0.275 $\mathrm{\mu m}$/pixel). The scanned images were manually reviewed to control for failed autofocus. The tumor slides of all the cases were reviewed by the pathologists (DS, YT, and IM) and the slides showing the deepest invasion into the bowel wall and/or the worst differentiated parts of the tumor, were selected for analysis.

\subsection*{Detection and classification of cells based on nuclear appearance}
Two separate convolutional neural networks (CNNs) were trained, one for detection and another for classification of cells \cite{sirinukunwattana2016locality}. A spatially-constrained CNN produced a probability map assigning to each pixel the probability of being the center of a cell. Subsequently, the locations of cells were estimated by the local maxima of the probability map. To classify a detected cell, multiple small sub-images in the neighborhood of the detected cell were extracted and then fed to the neighboring ensemble predictor (NEP). The NEP was trained to classify 4 cell types: malignant epithelial cells, inflammatory cells (including eosinophils, lymphocytes, and neutrophils), spindle-shaped cells (including normal fibroblasts, CAFs, and smooth muscle cells), and necrotic debris. 

The training and validation of the two algorithms were carried out on a dataset consisting of more than 20,000 cells, annotated by an experienced pathologist and a trained observer. The pixel resolution of images in the dataset was reduced to 0.55 $\mathrm{\mu m}$ (equivalent to using a $\times$20 microscope objective). This dataset consisted of certain H\&E-stained WSIs from cases that were initially excluded from the study. Based on a 2-fold cross-validation, the cell detection algorithm achieved an F1-score of 0.802 and the cell classification algorithm a multiclass AUC score \cite{hand2001simple} of 0.917. For more details of the cell detection and classification method, see Sirinukunwattana et al. \cite{sirinukunwattana2016locality}.

\subsection*{Quantifying local tissue characteristic}
We first split a WSI into small non-overlapping image tiles of size 200$\times$200 $\mathrm{\mu m^2}$ (Fig. \ref{fig:overview}a), which was within the limit of effective intercellular communication distance \cite{francis1997effective}. For each image tile, a cell network (in computational terms, a graph) was constructed based on cell detection and classification results (Fig. \ref{fig:overview}b). The vertices of the network represent cells of different types. The network itself is the associated Delaunay triangulation (Fig. \ref{fig:overview}c), so that an edge represents a connection between a pair of neighboring cells. The edges connecting cells in one tile with cells in an adjacent tile were not considered. Since there are 4 cell classes, there are 10 possible pairs of cell-cell connections in the network. We then used the distribution of different cell-cell connection types (Fig. \ref{fig:overview}d) to characterize a given image tile. 

\subsection*{Tissue phenotyping using cell-cell connection frequencies}
In order to group image tiles into different phenotypes, we first calculate a feature vector based on cell-cell connection frequencies. We consider the 4-element set $A=\{M,I,S,N\}$, where $M$ denotes the malignant epithelial type, $I$ the inflammatory type, $S$ the spindle-shaped type, and $N$ the necrotic debris type. We also identify $A$ with ${1,2,3,4}$ and define an indexing set $Q = \left\{(i,j) | i \geq j\right\}$. Let $h = \left[h_{(i,j)} | (i,j) \in Q\right] \in R^{10}$  be the ten-dimensional cell-cell connection frequency vector representing the frequencies of all cell-cell connections, where $h_{(i,j)} \in [0,1]$ denotes the proportion of connection frequencies between cells of types $i$ and $j$. We calculated this vector for every image tile extracted from every WSI in the dataset.

Next, we performed $k$-medoid clustering on all frequency vectors, calculated as above, for all tiles in all WSIs in the dataset in order to group image tiles into different phenotypes. This unsupervised algorithm (we used the $k$-medoid algorithm implemented in Matlab 2016b) automatically finds a set of medoids — representative frequency vectors for tile phenotypes within the data — and assigns a phenotype label to each tile according to its nearest medoid. We employed the Chi-squared distance between a frequency vector $h$ and a medoid $m$ given by:
\[
d(h,m) = \sum_{k\in Q} \frac{(h_k  - m_k)^2}{h_k + m_k}
\]
We initialized the medoids randomly and ran the clustering algorithm 100 times for each trial. We then used the results from the replicate that yielded the smallest total sum of distances between the frequency vectors and their corresponding medoids. The criteria used to determine the number of phenotypes $k$ were the similarity between the phenotypes and the correlation between tissue morphometric features derived from the phenotypes (described below). The similarity between a pair of phenotypes was measured in terms of the Chi-squared distance between the pair of medoids representing the phenotypes. Correlation between a pair of features was measured by the Spearman correlation coefficient. In order to find a suitable number of distinct phenotypes $k$, we chose the maximum number of phenotypes that produced relatively high values of Chi-squared distance and relatively low values of correlation between distinct features. A distance value less than 0.2 and a correlation coefficient value greater than 0.8 were considered undesirable. We found that $k$=6 is the maximum number of phenotypes that satisfies both criteria (Fig S4).

Examples of image tiles from different tissue phenotypes discovered using cell-cell connection frequencies are shown in Fig. \ref{fig:overview}e. As can be observed in Fig. \ref{fig:overview}e, the six connection frequency (CF) based phenotypes found automatically corresponded well with the following distinct tissue phenotypes: smooth muscle, inflammation, tumor-stroma interface, tumor, stroma, and necrosis. 

\subsection*{Tissue phenotyping based on appearance}

We also trained a deep learning based CNN for patch-based tissue phenotyping, in which the following 9 categories of image patches were explicitly considered: normal, non-tissue background, loose connective tissue (submucosa), fat (adipose), stroma (desmoplasia), inflammation, necrosis, smooth muscle, and tumor. Each image patch was of size 32$\times$32 pixels with a pixel resolution of 2.2 $\mathrm{\mu m}$/pixels (~5$\times$ objective). The architecture of the CNN was a simplified version of that proposed by Simonyan \textit{et al}. \cite{simonyan2014very}.

In developing this appearance (AP) based approach to tissue phenotyping, we used a dataset consisting of 193 sub-images, each of size 1,346$\times$982 pixels. These images were extracted from WSIs of cases that were initially excluded from the study. A trained observer (KS) annotated all images. We randomly split the images into three parts with 52.5\% for training, 17.5\% for validation, and 30\% for testing. Each WSI contributed images to only one part of the split. For training and validation, we extracted multiple patches of size 32$\times$32 pixels from the training and validation images. We selected the version of the algorithm that yielded the best performance on the validation part. In testing, for each test image, we extracted patches in a sliding-window fashion and classified each of them separately before merging the results together to obtain a segmentation result for the whole image. The correct classification accuracies for the 9 tissue phenotypes were as follows: normal 98.9\%, non-tissue background 99.9\%, loose connective tissue (submucosa) 98.4\%, fat (adipose) 97.9\%, stroma (desmoplasia) 90.4\%, inflammation 99.3\%, necrosis 98.2\%, smooth muscle 97.5\%, and tumor 96.0\%. 

We ran the trained segmentation algorithm on the 108 H\&E-stained WSI images, used in the analyses. Examples of the segmentation results can be seen in Fig. S1. Furthermore, as a quality control, segmentation results of 10 images (out of 108 images) were randomly selected and then reviewed by expert pathologists (DS, IC). 

\subsection*{Automatically-derived tissue phenotypic features}

The CF and AP based tissue phenotypic features were calculated as follows: 
\[ \text{phenotype ratio} =  \frac{\text{area of the tissue phenotype}}{\text{total tissue area}} 
\]
Here, the tissue area was computed from all tissue types excluding the normal and fat regions. The other tissue phenotypic features were quantified as follows:
\[ \text{stroma-tumor ratio} = \frac{\text{stroma area}}{\text{stroma area} + \text{tumor area}}
\] 
\[ \text{necrosis-tumor ratio} = \frac{\text{necrosis area}}{\text{necrosis area} + \text{tumor area}}
\] 
where stroma, tumor, and necrosis areas were obtained from the AP based phenotyping results. 

\subsection*{Statistical analyses}
Our analysis did not distinguish well differentiated from moderately differentiated tumors—as recommended by Compton et al. \cite{compton2000prognostic,compton2000updated}, this helps to avoid contradictory labelling by two different observers, or even by a single observer, looking at the same sample on two different occasions. Missing data were filled in with 100 imputed values using the multiple imputation method implemented in the R `mice' library \cite{buuren2011mice}. Analyses were performed on every imputed dataset and the results were combined to yield an overall estimate \cite{rubin2004multiple}. The significance level was set to 0.05 for all the tests described below.

Association between the tissue phenotypic and standard clinical features was tested by the Mann-Whitney test and the strength of association was determined through coefficients of determination ($r^2$) of the test \cite{cohen2008explaining,fritz2012effect}. The median p-value and $r^2$ were reported for a variable with multiple imputed values. We used the `rms' library in R \cite{harrell2016rms} to fit logistic regression models, to calculate the area under the receiver operating characteristic curve (AUC), and to perform survival and bootstrap analyses.

Logistic regression analysis was performed to assess the predictive power of each phenotypic feature in identifying patients with concurrent distant metastasis at the time of diagnosis or a propensity for distant metastasis development. Effects of the automatically-derived features were gauged after adjusting for the standard clinical variables in multivariate logistic regression models. A total of 108 cases (78 UHCW and 30 HGH) were used in the analysis. The 5-year metastasis status was treated as a binary outcome and features were treated as predictors in regression models. Estimated odds ratio and its 95\% CI were obtained for each feature to quantify the risk of distant metastasis development associated with the feature. We reported the factor of change in odds ratio when the value of a feature changes from the baseline value to the new value. For a continuous feature, the baseline and the changed values were set to the 1st and 3rd quartiles of the feature. Furthermore, likelihood ratio p-values were computed to assess goodness of fit of predictive models contributed by various features. 

Survival analysis was performed to determine the prognostic value for DMFS associated with each feature. Univariate and multivariate Cox proportional hazards regression analyses were conducted on 78 cases from the UHCW cohort for which DMFS data were available. The former was used to evaluate the prognostic impact of each feature separately while the latter was used to assess the prognostic value of image-based tissue phenotypic features while adjusting for the effects of the clinical features. Rao's score test and Wald test were employed in the univariate and multivariate analyses, respectively, to test whether the regression coefficient corresponding to a particular feature in the Cox proportional hazards model was nonzero. Note that the score test is equivalent to the log-rank test when only a single categorical feature is considered in the model \cite{therneau2000modeling}. Hazard ratio and 95\% CI estimates were obtained for each feature. To internally validate the performance of each fitted Cox proportional hazards model in predicting the survival probability, a bootstrap routine \cite{harrell2015regression} with 100 resampling replicates was employed to estimate the AUC. The statistical significance difference between survival stratifications was determined through the log-rank test using the R `survival' library \cite{therneau2015package}. The cutoff with minimum p-value was used for stratification, and the p-value was adjusted according to Altman's correction \cite{altman1994dangers} in case of a continuous feature. 

\subsection*{Data Availability}
The datasets generated during and/or analyzed during the current study are available from the corresponding author on reasonable request. Extracted image features and codes to perform statistical analyses will be included in the Supplementary Information files once the manuscript has been accepted for publication.
% The whole slide images and associated clinical datasets generated during and/or analysed during the current study are available from the corresponding author on reasonable request. Extracted image features and codes to perform statistical analyses are included in this published article (and its Supplementary Information files).

\bibliography{sample}

\section*{Acknowledgements}
This paper was made possible by NPRP grant number NPRP5-1345-1-228 from the Qatar National Research Fund (a member of Qatar Foundation). The statements made herein are solely the responsibility of the authors. The authors would like to acknowledge the contribution by Sean James, Kayleigh Patterson, Dr. Aisha Meskiri, and Dr. Asha Rupani who involved in the preparation, staining, and scanning of histology slides of the CRC samples used in this study. 

\section*{Author contributions statement}
NR, DS, IC, and KS designed the study. ZA and DS collected the clinical data. IM and DS reviewed and graded the histological samples. DS, YT, and IM reviewed and selected slides for analysis. KS and NR developed the image analysis tools. YT and KS generated ground truth data for training the nuclear detection and classification algorithm. YT, DS, and IC reviewed the results of the algorithms. KS conducted all the statistical analyses. NR, DE, and IC supervised the statistical analyses and interpreted the results. KS drafted the manuscript. All authors were involved in discussion of the results and finalization of the manuscript.

\section*{Additional information}

\textbf{Competing financial interests:} The author(s) declare no competing financial interests.

\begin{table}[ht]
\centering
\begin{tabular}{|l|c|c|c|}
\hline
Clinical feature                                                                 & UHCW cohort           & HGH cohort            & Total                 \\ \hline
Number of cases                                                                  & 78                    & 30                    & 108                   \\ \hline
Tumor histological type                                                          &                       &                       &                       \\ 
\hspace{0.5cm}Adenocarcinoma                                                                   & 69                    & 26                    & 95                    \\ 
\hspace{0.5cm}Mucinous                                                                         & 9                     & 3                     & 12                    \\ 
\hspace{0.5cm}Not available                                                                    & 0                     & 1                     & 1                     \\ \hline
Tumor differentiation                                                            &                       &                       &                       \\ 
\hspace{0.5cm}Well differentiated                                                              & 6                     & 6                     & 12                    \\ 
\hspace{0.5cm}Moderately differentiated                                                        & 39                    & 20                    & 59                    \\ 
\hspace{0.5cm}Poorly differentiated                                                            & 17                    & 4                     & 21                    \\ 
\hspace{0.5cm}Not available                                                                    & 16                    & 0                     & 16                    \\ \hline
T stage                                                                          &                       &                       &                       \\ 
\hspace{0.5cm}pT3                                                                              & 60                    & 26                    & 86                    \\ 
\hspace{0.5cm}pT4                                                                              & 18                    & 4                     & 22                    \\ \hline
5-year metastasis                                                                &                       &                       &                       \\ 
\hspace{0.5cm}No                                                                               & 52                    & 23                    & 75                    \\ 
\hspace{0.5cm}Yes                                                                              & 26                    & 7                     & 33                    \\ \hline
\begin{tabular}[c]{@{}l@{}}Median metastasis-free\\ survival (year)\end{tabular} & \multicolumn{1}{l|}{} & \multicolumn{1}{l|}{} & \multicolumn{1}{l|}{} \\ 
\hspace{0.5cm}With distant metastasis                                                          & 1.0410                & Not available         & 1.0410                \\ 
\hspace{0.5cm}Without distant metastasis                                                       & \textgreater 5        & Not available         & \textgreater 5        \\ \hline
\end{tabular}
\caption{\label{tab:clinical_data} A summary of clinical data.}
\end{table}

\begin{table}[t]
\setlength{\tabcolsep}{3pt}
\centering
\begin{tabular}{lcccccc}
\hline
\multicolumn{1}{c}{\multirow{2}{*}{Feature}} & \multicolumn{2}{c}{Differentiation} & \multicolumn{2}{c}{Histological type} & \multicolumn{2}{c}{T stage}  \\ \cline{2-7} 
\multicolumn{1}{c}{}                         & p-value     & $r^2$    & p-value      & $r^2$     & p-value & $r^2$ \\ \hline
{\footnotesize CF smooth muscle ratio}                         & 0.711       & 0.001                  & 0.567        & 0.003                   & 0.834   & 0                   \\ \hline
{\footnotesize CF inflammation ratio}                          & 0.168       & 0.018                  & 0.957        & 0                       & 0.108   & 0.024               \\ \hline
\begin{tabular}[c]{@{}l@{}}{\footnotesize CF tumor-stroma}\\ {\footnotesize interface ratio}\end{tabular}                 & 0.339       & 0.008                  & 0.608        & 0.002                   & \textbf{0.002}   & 0.092               \\ \hline
{\footnotesize CF tumor ratio}                                 & \textbf{0.033}       & 0.042                  & 0.587        & 0.003                   & 0.466   & 0.005               \\ \hline
{\footnotesize CF stroma ratio}                                & 0.655       & 0.002                  & 0.949        & 0                       & 0.362   & 0.008               \\ \hline
{\footnotesize CF necrosis ratio}                              & 0.321       & 0.009                  & 0.267        & 0.011                   & 0.973   & 0                   \\ \hline
\end{tabular}
\caption{\label{tab:association} Association between the CF tissue phenotypic features and standard clinical features. Mann-Whitney test's p-value and coefficient of determination ($r^2$) are used to assess the association between features. The results with p-value less than 0.05 is considered statistically significant (bold).}
\end{table}

\begin{table*}[t]
\centering

\begin{tabular}{lcccccc}
\hline
\multicolumn{1}{c}{\multirow{2}{*}{Feature}} & \multicolumn{3}{c}{Univariate}                                                & \multicolumn{3}{c}{Multivariate}                                              \\ \cline{2-7} 
\multicolumn{1}{c}{}                                                                                   & Odds ratio factor                                             & p-value & AUC       & Odds ratio factor                                             & p-value & AUC        \\ \hline
\multicolumn{7}{c}{Standard histological features}                                                                                                                                                                                                                       \\ 
Differentiation {\scriptsize (MD $\rightarrow$ PD)}                                         & 1.729 {\scriptsize (0.622,4.807)} & 0.298    &  0.525    & \multicolumn{3}{l}{}                                                          \\ 
Histological type {\scriptsize (Adenocarcinoma $\rightarrow$ Mucinous)}                      & 0.752 {\scriptsize (0.190,2.984)} & 0.679    &  0.491    & \multicolumn{3}{l}{}                                                          \\ 
T stage {\scriptsize	(pT3 $\rightarrow$ pT4)}                                                & 2.283 {\scriptsize (0.869,5.995)} & 0.097  &  0.559      & \multicolumn{3}{l}{}                                                          \\ \hline
\multicolumn{7}{c}{Connection frequency (CF) based tissue phenotypic features}                                                                                                                                                                                           \\ 
CF smooth muscle ratio {\scriptsize (0.167 $\rightarrow$ 0.372)}                             & 2.258 {\scriptsize (1.117,4.565)} & \textbf{0.010} &  0.641 & 2.350 {\scriptsize (1.132,4.876)} & \textbf{0.008} & 0.638 \\ 
CF inflammation ratio {\scriptsize (0.042 $\rightarrow$ 0.138)}                              & 0.279  {\scriptsize (0.119,0.656)} & \textbf{0.009} & 0.661  & 0.307 {\scriptsize (0.126,0.749)} & \textbf{0.022} & 0.623\\ 
CF tumor-stroma interface ratio {\scriptsize (0.122 $\rightarrow$ 0.229)}                    & 1.326 {\scriptsize (0.722,2.435)} & 0.644        & 0.503 & 1.177  {\scriptsize (0.623,2.224)} & 0.800  & 0.556 \\ 
CF tumor ratio {\scriptsize (0.079 $\rightarrow$ 0.229)}                                     & 0.488 {\scriptsize (0.244,0.976)} & 0.111       &  0.623 & 0.520  {\scriptsize (0.252,1.070)} & 0.187  & 0.607 \\ 
CF stroma ratio {\scriptsize (0.174 $\rightarrow$ 0.275)}                                    & 0.567 {\scriptsize (0.334,0.962)} & 0.094      &  0.595  & 0.585 {\scriptsize (0.340,1.007)} & 0.139  &  0.580      \\ 
CF necrosis ratio {\scriptsize (0.022 $\rightarrow$ 0.053)}                                  & 0.713 {\scriptsize (0.366,1.386)} & 0.284      &  0.539  & 0.738  {\scriptsize (0.371,1.469)} & 0.254  &   0.539     \\ \hline
\multicolumn{7}{c}{Appearance (AP) based tissue phenotypic features}                                                                                                                                                                                                     \\
AP smooth muscle ratio {\scriptsize (0.136 $\rightarrow$ 0.330)}                             & 1.449 {\scriptsize (0.755,2.779)} & 0.513    & 0.511     & 2.075 {\scriptsize (0.970,4.437)} & 0.125  &  0.562      \\ 
AP inflammation ratio {\scriptsize (0.025 $\rightarrow$ 0.070)}                             & 0.422 {\scriptsize (0.206,0.865)} & \textbf{0.027} & 0.618 & 0.472 {\scriptsize (0.225,0.987)} & 0.056 &  0.585  \\ \hline
\multicolumn{7}{c}{Other features}                                                                                                                                                                                                                                       \\
Morisita index \cite{maley2015ecological} {\scriptsize (0.344 $\rightarrow$ 0.529)}                                & 0.973 {\scriptsize (0.533,1.775)} & 0.916      &  0.494  & 0.868 {\scriptsize(0.461,1.632)} & 0.866 &  0.541      \\ 
Stroma-tumor ratio \cite{mesker2007carcinoma, west2010proportion, huijbers2012proportion} {\scriptsize (0.400 $\rightarrow$ 0.613)}    & 0.898 {\scriptsize (0.515,1.567)} & 0.284      &  0.571  & 0.852 {\scriptsize (0.481,1.509)} & 0.329 &   0.569      \\ 
Necrosis-tumor ratio \cite{jayasinghe2015histopathological, pollheimer2010tumor, richards2012prognostic} {\scriptsize (0.077 $\rightarrow$ 0.224)}                  & 0.556 {\scriptsize (0.274,1.129)} & 0.247       & 0.551  & 0.577 {\scriptsize (0.278,1.197)} & 0.272 &    0.578     \\ \hline
\end{tabular}

\caption{\label{tab:logistic} Prognostic values of different features according to the logistic regression analysis. Each morphological feature is adjusted by the standard histological features in the multivariate analysis. The statistical significance of each feature is assessed by the likelihood ratio test's p-value. An interquartile change for a continuous variable or categorical change for a categorical variable is noted by ($x \rightarrow y$). A 95\% confidence interval of the estimate of odds ratio factor is noted by ($x$,$y$). A statistically significant result at the 0.05 is highlighted in bold. AUC in the multivariate analysis refers to the AUC of the multivariate model rather than an individual feature.}

\end{table*}

\begin{table*}[t]
\centering
\scalebox{0.8}{
\begin{tabular}{lccccccl}
\hline
\multicolumn{1}{c}{\multirow{2}{*}{Feature}} & \multicolumn{4}{c}{Univariate}                                                                                                                                                 & \multicolumn{3}{c}{Multivariate}                                                         \\ \cline{2-8} 
\multicolumn{1}{c}{}                                                                                   & \begin{tabular}[c]{@{}c@{}}Hazard ratio\\factor\end{tabular}   & \begin{tabular}[c]{@{}c@{}}Score test\\ p-value\end{tabular} & \begin{tabular}[c]{@{}c@{}}Log-rank test\\ p-value\end{tabular} & \multicolumn{1}{l}{AUC} & \begin{tabular}[c]{@{}c@{}}Hazard ratio\\factor\end{tabular} & \begin{tabular}[c]{@{}c@{}}Wald test\\ p-value\end{tabular} & AUC   \\ \hline
\multicolumn{8}{c}{Standard histological features}                                                                                                                                                                                                                                                                                                                                   \\ 
Differentiation {\scriptsize (MD $\rightarrow$ PD)}                                                                                  & 1.275 {\scriptsize (0.496,3.281)} & 0.539                                                        & 0.539                                                           & 0.496                    & \multicolumn{3}{l}{}                                                                     \\ 
Histological type {\scriptsize (Adenocarcinoma $\rightarrow$ Mucinous)}                                                              & 0.580 {\scriptsize (0.137,2.449)} & 0.450                                                        & 0.450                                                           & 0.512                    & \multicolumn{3}{l}{}                                                                     \\ 
T stage {\scriptsize (pT3 $\rightarrow$ pT4)}                                                                                        & 1.814 {\scriptsize (0.787,4.181)} & 0.156                                                        & 0.156                                                           & 0.542                    & \multicolumn{3}{l}{}                                                                     \\ \hline
\multicolumn{8}{c}{Connection frequency (CF) based tissue phenotypic features}                                                                                                                                                                                                                                                                                                       \\ 
CF smooth muscle ratio {\scriptsize (0.167 $\rightarrow$ 0.372)}                                                                     & 2.138 {\scriptsize (0.922,4.955)} & \textbf{0.016}                                               & \textbf{0.036}                                                  & 0.633                    & 2.467 {\scriptsize (1.062,5.730)}  & \textbf{0.008}                                              & 0.623 \\ 
CF inflammation ratio {\scriptsize (0.042 $\rightarrow$ 0.138)}                                                                      & 0.412 {\scriptsize (0.204,0.834)} & \textbf{0.039}                                               & 0.258                                                           & 0.608                    & 0.413 {\scriptsize (0.201,0.849)} & 0.051                                                       & 0.582 \\ 
CF tumor-stroma interface ratio {\scriptsize (0.122 $\rightarrow$ 0.229)}                                                            & 1.120 {\scriptsize (0.619,2.029)} & 0.931                                                        & 0.445                                                           & 0.462                    & 1.054 {\scriptsize (0.575,1.934)} & 0.968                                                       & 0.530 \\ 
CF tumor ratio {\scriptsize (0.079 $\rightarrow$ 0.229)}                                                                             & 0.584 {\scriptsize (0.303,1.125)} & 0.174                                                        & 0.399                                                           & 0.584                    & 0.520 {\scriptsize (0.304,1.184)} & 0.276                                                       & 0.566 \\ 
CF stroma ratio {\scriptsize (0.174 $\rightarrow$ 0.275)}                                                                            & 0.475 {\scriptsize (0.252,0.896)} & \textbf{0.032}                                               & \textbf{0.047}                                                  & 0.627                    & 0.469 {\scriptsize (0.245,0.895)} & 0.062                                                       & 0.616 \\ 
CF necrosis ratio {\scriptsize (0.022 $\rightarrow$ 0.053) }                                                                          & 0.689 {\scriptsize (0.343,1.386)} & 0.233                                                        & 0.595                                                           & 0.535                    & 0.619 {\scriptsize (0.302,1.267)} & 0.202                                                       & 0.552 \\ \hline
\multicolumn{8}{c}{Appearance (AP) based tissue phenotypic features}                                                                                                                                                                                                                                                                                                                 \\ 
AP smooth muscle ratio {\scriptsize (0.136 $\rightarrow$ 0.330)}                                                                     & 1.515 {\scriptsize (0.762,3.010)} & 0.467                                                        & 0.759                                                           & 0.514                    & 2.086 {\scriptsize (1.025,4.247)} & 0.127                                                       & 0.563 \\ 
AP inflammation ratio {\scriptsize (0.025 $\rightarrow$ 0.070)}                                                                      & 0.401 {\scriptsize (0.216,0.742)} & \textbf{0.001}                                               & \textbf{0.003}                                                  & 0.654                    & 0.431 {\scriptsize (0.224,0.832)} & \textbf{0.009}                                              & 0.625 \\ \hline
\multicolumn{8}{c}{Other features}                                                                                                                                                                                                                                                                                                                                                   \\ 
Morisita index \cite{maley2015ecological} {\scriptsize (0.344 $\rightarrow$ 0.529)}                                                                        & 0.844 {\scriptsize (0.458,1.556)} & 0.861                                                        & 0.795                                                           & 0.507                    & 0.742 {\scriptsize (0.397,1.387)} & 0.646                                                       & 0.502 \\ 
Stroma-tumor ratio \cite{mesker2007carcinoma, west2010proportion, huijbers2012proportion} {\scriptsize (0.400 $\rightarrow$ 0.613)}                                                             & 0.880 {\scriptsize (0.517,1.498)} & 0.607                                                        & 0.529                                                           & 0.519                    & 0.813 {\scriptsize (0.464,1.424)} & 0.583                                                       & 0.515 \\ 
Necrosis-tumor ratio \cite{jayasinghe2015histopathological, pollheimer2010tumor, richards2012prognostic} {\scriptsize (0.077 $\rightarrow$ 0.224)}                                                          & 0.647 {\scriptsize (0.328,1.275)} & 0.281                                                        & 0.370                                                           & 0.554                    & 0.652 {\scriptsize (0.331,1.286)} & 0.216                                                       & 0.575 \\ \hline
\end{tabular}}
\caption{\label{tab:cox} Prognostic values of different features according to the Cox proportional hazards regression analysis on the UHCW cohort. Each morphological feature is adjusted by the standard histological features in the multivariate analysis. An interquartile change for a continuous variable or categorical change for a categorical variable is noted by ($x \rightarrow y$). A 95\% confidence interval of the estimate of hazard ratio factor is noted by ($x$,$y$). A statistically significant result at the 0.05 significance level is highlighted in bold. AUC in the multivariate analysis refers to the AUC of the multivariate model rather than an individual feature.}
\end{table*}

\end{document}